\documentclass[letterpaper, 10 pt, conference]{ieeeconf}  

\IEEEoverridecommandlockouts                              

\usepackage{graphicx}
\usepackage{soul, color, xcolor}
\usepackage{amsmath}
\usepackage{amsfonts}
\usepackage{booktabs}
\usepackage{multirow}

\usepackage{graphicx}
\usepackage{wrapfig}
\usepackage{stfloats}
\usepackage{caption}
\usepackage{longtable}
\usepackage{subfigure}
\usepackage{hyperref}

\title{\LARGE \bf
VideoSAM: Open-World Video Segmentation
}

\author{
Pinxue Guo$^{1,*}$, 
Zixu Zhao$^{2,*}$,
Jianxiong Gao$^3$,
Chongruo Wu$^2$,
Tong He$^2$, 
Zheng Zhang$^2$, \\
Tianjun Xiao$^{2,\dagger}$,
Wenqiang Zhang$^{1,\dagger}$ 
}

\begin{document}

\twocolumn[{
\renewcommand\twocolumn[1][]{#1}
\maketitle
\begin{center}
    \vspace{-8mm}
    \centering
    \captionsetup{type=figure}
    \includegraphics[width=1.0\textwidth]{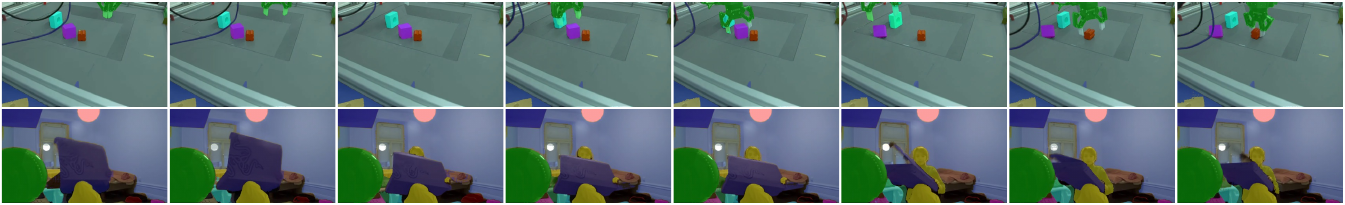}
    \vspace{-5mm}
    \captionof{figure}{\small VideoSAM produces open-world segmentation on videos with consistent object granularity. One color indicates one object.}
    \label{fig:teaser}
\end{center}
}]

\begingroup
  \renewcommand{\thefootnote}{}
  \footnotetext{$^*$ Equal Contribution, $^\dagger$ Corresponding authors.}
  \footnotetext{$^1$ Academy for Engineering and Technology, Fudan University. $^2$ Amazon Web Services Shanghai AI Lab. $^3$ Institute of Science and Technology for Brain-inspired Intelligence, Fudan University.}
\endgroup

\thispagestyle{empty}
\pagestyle{empty}

\begin{abstract}


Video segmentation is essential for advancing robotics and autonomous driving, particularly in open-world settings where continuous perception and object association across video frames are critical. While the Segment Anything Model (SAM) has excelled in static image segmentation, extending its capabilities to video segmentation poses significant challenges. We tackle two major hurdles: a) SAM’s embedding limitations in associating objects across frames, and b) granularity inconsistencies in object segmentation.  To this end, we introduce VideoSAM, an end-to-end framework designed to address these challenges by improving object tracking and segmentation consistency in dynamic environments. VideoSAM integrates an agglomerated backbone, RADIO, enabling object association through similarity metrics and introduces Cycle-ack-Pairs Propagation with a memory mechanism for stable object tracking. Additionally, we incorporate an autoregressive object-token mechanism within the SAM decoder to maintain consistent granularity across frames. Our method is extensively evaluated on the UVO and BURST benchmarks, and robotic videos from RoboTAP, demonstrating its effectiveness and robustness in real-world scenarios. All codes will be available.
\end{abstract}
\section{INTRODUCTION}
Video segmentation is crucial in the development of robotics~\cite{hurtado2022semantic, 9561690} and autonomous driving~\cite{siam2021video, muhammad2022vision}, presenting unparalleled challenges in open-world scenarios. The ability to accurately segment and associate anything across video frames is foundational for understanding dynamic environments, essential for safe and efficient navigation~\cite{rashed2019motion} and manipulation~\cite{9664632}. 
The Segment Anything Model (SAM)~\cite{kirillov2023segment}, has shown remarkable success in open-world image segmentation. Its ability to segment any object in an image without specific prior knowledge highlights its potential for addressing open-world challenges. However, applying SAM to videos introduces many challenges, especially in robotics where real-world environments require continuous perception over time.
Recent efforts~\cite{cheng2023segment, rajivc2023segment,yang2023track} seek to integrate SAM with existing propagation models~\cite{yang2022decoupling, cheng2022xmem,karaev2023cotracker}. While useful, these approaches lack end-to-end optimization and depend on human-defined inductive biases or rules that are less effective in complex scenarios.

\begin{figure}[t]
  \centering
  \includegraphics[width=0.96\linewidth]{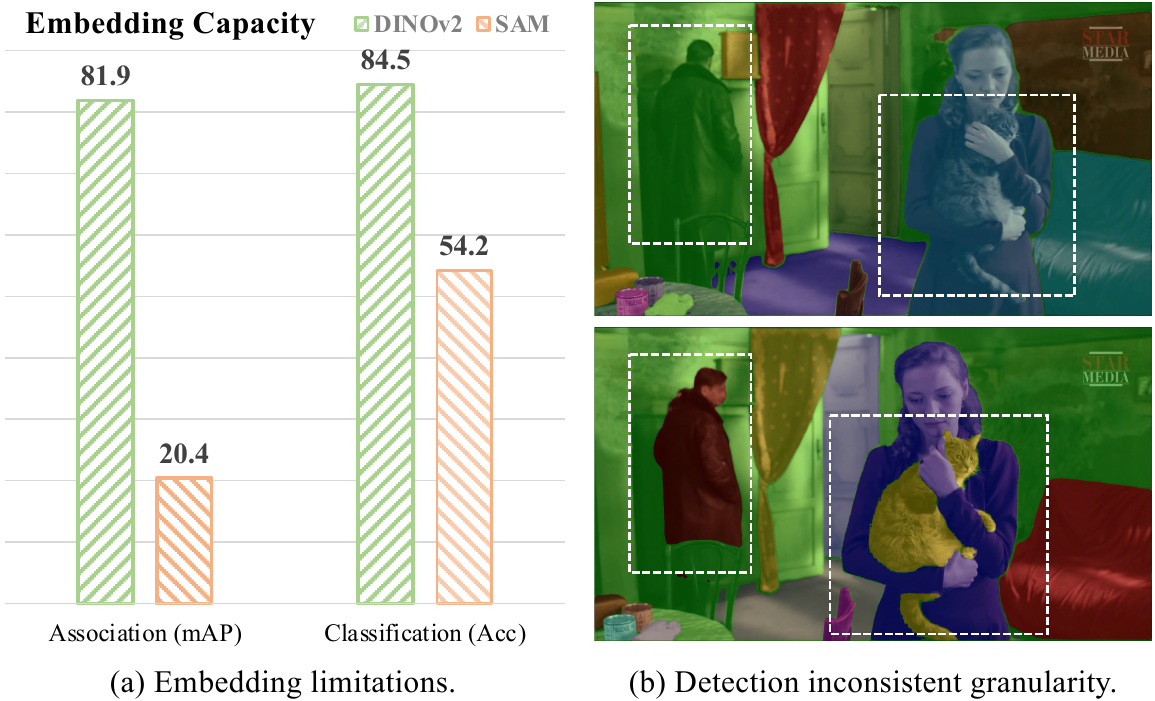}
  \caption{\small Limitations of the Segment Anything Model (SAM) in extending to the open-world video segmentation task. (a) We conduct oracle experiments using ground-truth masks for each frame to associate objects across frames, utilizing DINOv2/SAM embeddings pooled from the masks on the OVIS~\cite{qi2022occluded}. We also perform linear probing experiments on ImageNet~\cite{deng2009imagenet}. These experiments demonstrate that while SAM embeddings are powerful for static image segmentation, they lack association and semantic information. (b) SAM exhibits inconsistent granularity when detecting objects across different frames, e.g., woman and cat.}
  \label{fig:motivation}
  \vspace{-3mm}
\end{figure}

In this paper, we identify two major challenges in extending SAM to video scenarios, as shown in Fig.~\ref{fig:motivation}.
1) \textit{SAM embedding limitations}: 
SAM mask embeddings, although effective for segmentation, struggle with associating objects across video frames. This is because SAM is highly specialized for segmentation, relying predominantly on high-level features from the MAE~\cite{he2022masked}-ViT~\cite{dosovitskiy2020image} backbone, which limits its capacity to assess visual similarities between frames
2) \textit{Granularity inconsistency across frames}: SAM often produces inconsistent segmentation granularity across frames, even prompted at the same object level. This inconsistency disrupts continuous object tracking and hinders the model’s ability to provide stable and reliable segmentation in videos.

To tackle these challenges, we propose VideoSAM, an end-to-end framework designed to consistently track and segment objects across frames while maintaining coherent granularity. 
First, we integrate an agglomerated backbone, named RADIO~\cite{ranzinger2023radio}, to overcome SAM embedding’s limitations in association. RADIO can perform well for both segmentation and visual similarity assessment, laying the foundation for a tracking system based on similarity metrics. Then, we introduce Cycle-ack Pairs Propagation and its corresponding memory mechanism in our system, designed to ensure high-confidence tracking of objects across frames, maintaining accurate and stable object identities throughout the tracking process.
Furthermore, to resolve the inconsistency in granularity, we introduce the AR-SAM Decode with autoregressive object prompt mechanism, adapted from SAM mask decoder. This object prompt is designed to carry forward object-level information across frames, ensuring stable and coherent segmentation. Meanwhile, the corresponding cycle-consistent training encourages consistent object granularity throughout the sequence, further enhancing robustness in dynamic scenes.
These not only achieve segmenting and tracking object with the consistent granularity across video frames in open-world scenario but also make the model end-to-end differentiable, enabling data-driven optimization to address tracking issues, compared to existing approaches relying on human-defined inductive biases that are less effective for complex scenarios. We validate the open-world video segmentation capabilities of the VideoSAM framework through extensive experiments on the UVO~~\cite{wang2021unidentified} and BURST~\cite{athar2023burst} benchmarks. Additionally, we apply VideoSAM to real-world robotic videos from RoboTAP~\cite{vecerik2024robotap}, further evaluating its general applicability across diverse environments.
We summarize our contributions as follows:
\begin{itemize}
    \item We introduce VideoSAM, an end-to-end framework that extends SAM to open-world video segmentation, specifically addressing the challenges of object association and consistent granularity across frames.

    \item We integrate additional visual embeddings and introduce Cycle-ack Pairs Propagation with a memory mechanism to achieve efficient, robost, and stable object tracking across frames.

    \item We present the AR-SAM decoder, incorporating the autoregressive object prompt within the SAM mask decoder. Combined with the cycle-sequential training strategy, it effectively addresses the issue of inconsistent segmentation granularity by differentiable optimization across frames.

    \item We demonstrate the general applicability of VideoSAM through extensive experiments on open-world benchmarks, also validating its potential in real-world robotic applications. 

\end{itemize}

\vspace{-1mm}
\section{Related Works}

\subsection{Open-world segmentation}
Open-world segmentation~\cite{wang2021unidentified, qi2022open}, a cutting-edge area in computer vision, challenges the norm by aiming to identify and delineate objects beyond pre-defined classes, crucial for applications like autonomous driving and robotic manipulation. It builds on advancements in open-world learning, such as incremental learning~\cite{wu2019large}, few-shot learning~\cite{parnami2022learning}, and meta-learning~\cite{hsu2018unsupervised,9561690}, to tackle catastrophic forgetting and adaptively incorporate new object classes. Adaptations of semantic segmentation models for open-world settings focus on recognizing and categorizing "unknown" entities through self-supervised learning and anomaly detection. 
Recently, the Segment Anything model (SAM)~\cite{kirillov2023segment} has demonstrated strong zero-shot performance on image-level segmentation. The capability is gained through a data-engine that has a broader coverage of objects on scale and diversity. This effort has provide possibilities to more perception tasks. However, feature representation from SAM is proven not strong enough to support tasks requires strong semantic and association capabilities.

\vspace{-2mm}
\subsection{Object tracking}
Object tracking is a video-level task has vast application in video analysis~\cite{chen2023seqtrack,Zhao_2023_ICCV}, robotics~\cite{9811873,gad2022multiple} and autonomous driving~\cite{luo2021exploring}. Usually, there are two streams of tracking pipeline: propagation-based and detection-based. Propagation-based methods propagate object mask from one frame to the next utilizing position, motion, similarity signals. Typical propagation-based tracking methods include XMem~\cite{cheng2022xmem}, AOT~\cite{yang2021associating}, DeAOT~\cite{yang2022decoupling}, etc. Detection-based tracking first run object detector on each frame, then associate object in different frames using feature similarity. Typical detection-based tracking approaches include MOTR~\cite{zeng2022motr}, TransTrack~\cite{sun2020transtrack}, OC-MOT~\cite{Zhao_2023_ICCV}, etc. Occlusion is a common challenge for tracking. Memory is proven to be a useful architecture to mitigate this issue. XMem provides a memory solution for propagation-based tracking. MeMOT~\cite{cai2022memot} provides that for detection-based tracking. OC-MOT provides the self-supervised way to build tracking memory.

\vspace{-2mm}
\subsection{Existing tracking-anything model}
Though SAM is an image-level model, several efforts started to integrate SAM with existing tracking modules in a plug-and-play way. TAM~\cite{yang2023track} integrates SAM with the propagation model X-mem~\cite{cheng2022xmem}. SAM-Track~\cite{cheng2023segment} integrates SAM with DeAOT~\cite{yang2022decoupling}. The model DEVA~\cite{cheng2023tracking} didn't use SAM in their paper submission but could be easily integrated. SAM-PT~\cite{rajivc2023segment} integrates SAM with a point-tracker, such as Cotracker~\cite{karaev2023cotracker}, propagating position prompts into the next frames. However, these methods lack end-to-end optimization and rely on human-defined inductive biases or rules that are less effective in complex scenarios. And SAM2 is a concurrent work with a focus on interactive segmentation but will suffer from high computational and memory costs in automatic video segmentation scenario, due to its per-object correspondence and memory design.

\section{Methods}

\begin{figure*}[ht]
  \centering
  \includegraphics[width=1.0\linewidth]{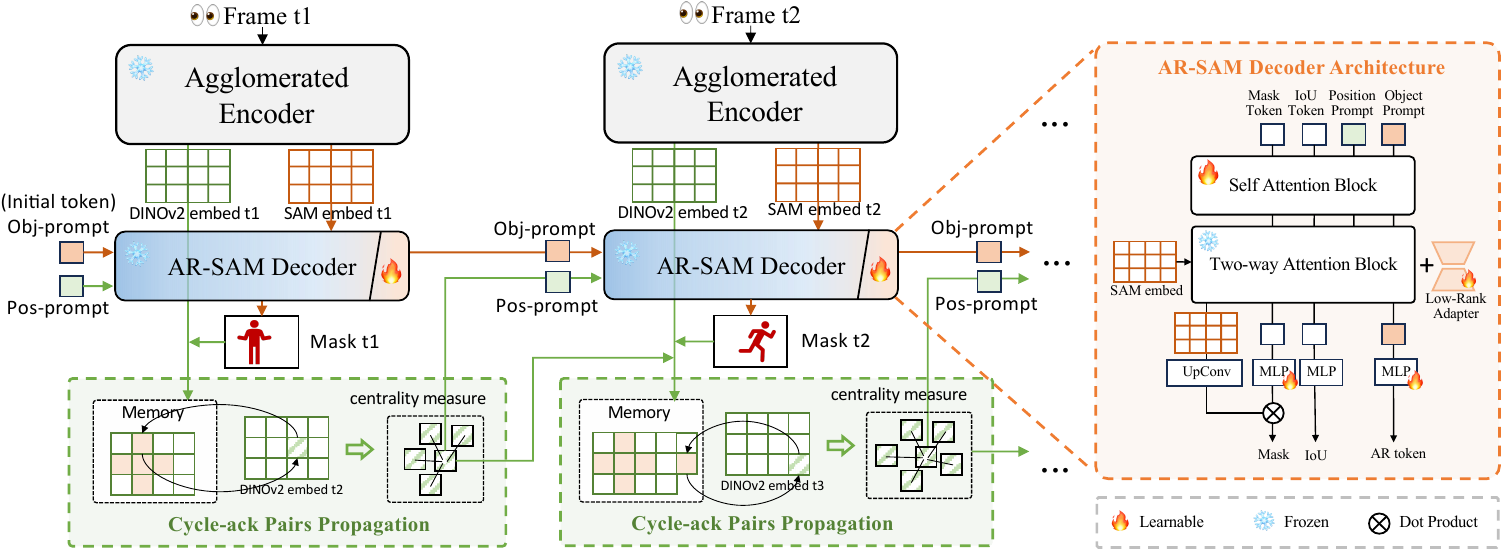}
  \caption{Overview of VideoSAM. It simultaneously employs DINOv2 and SAM embeddings from an efficient agglomerated encoder to handle object association and segmentation, respectively. Cycle-ack Pair Propagation is introduced to robustly associate objects across frames. AR-SAM Decoder, adapted from the mask decoder of SAM with temporal autoregressive object prompts, is used to maintain consistent segmentation granularity across video frames.}
  \label{fig:overview}
  \vspace{-4mm}
\end{figure*}

In this section, we introduce the architecture and components of the proposed VideoSAM framework, detailing how it extends SAM to the video domain for automatic open-world segmentation. We first provide an overview of SAM to establish the foundation for VideoSAM, followed by an in-depth discussion of the robust object association and granularity consistent segmentation we developed to address the key challenges of object association and granularity consistency across video frames.

\subsection{Preliminaries of SAM}
The Segment Anything Model (SAM)~\cite{kirillov2023segment} is a recent vision foundation model achieving state-of-the-art open-set image segmentation. It has showcased robust generation to produce high-quality masks from prompts including points, boxes, and masks. SAM consists of an image encoder, a flexible prompt encoder, and a lightweight mask decoder. The image encoder is a Vision Transformer (ViT) that transforms the high-resolution images to an image embedding of $h\times w \times c$ spatial size. The prompt encoder projects prompts into $c$-dimensional tokens as the input of the decoder. Besides prompt tokens, SAM decoder also takes three mask tokens as input, each indicating the mask granularity from subpart, part to whole object, and an extra IoU token indicating the mask reliability. The mask decoder then integrates these tokens with image embeddings to generate masks corresponding to three levels of granularity. 

However, in the context of detection-based tracking, we identify two issues of SAM: 1) \textit{Limited tracking capability of SAM image embeddings.} SAM image encoder are trained on large-scale SA-1B dataset but with segmentation loss only, which attributes to the fact that SAM image embeddings mainly preserve the localization signals for segmentation, while losing the crucial visual information such as semantics and appearances that are necessary for video correspondence learning. In our experiments as shown in Fig.~\ref{fig:motivation}(a), SAM image embeddings have demonstrated poor performance over the classification task under linear probe, and detection-based tracking task under mask feature pooling.
2) \textit{Granularity inconsistency of SAM masks.} Fig.~\ref{fig:motivation}(b) illustrates the granularity inconsistency issue of SAM across frames. SAM’s mask tokens are unable to maintain a specific segmentation granularity, making it challenging to achieve consistent object tracking, such as “always segmenting the complete person,” even when only using the object-level granularity.

\subsection{VideoSAM Framework}
To address the challenges of extending SAM from static images to videos, we propose VideoSAM, an end-to-end framework designed for open-world video segmentation. 
VideoSAM extends SAM by introducing components for continuous and robust object tracking while maintaining consistent granularity across video frames. Specifically, it associates objects via tracking position prompts through the proposed Cycle-ack Pairs Propagation and segments them using the AR-SAM Decoder with a temporal autoregressive object prompt, as shown in Fig~\ref{fig:overview}. In this framework, we fully exploit and preserve the open-world capabilities of the foundation model at each step, leveraging DINOv2 all-purpose embeddings for effective object association and SAM embeddings and decoder for zero-shot segmentation.
This framework preserves SAM’s strong segmentation ability while overcoming the two significant issues mentioned above. Additionally, instead of following tracking-by-detection paradigm, it avoids the inefficiency and uncorrectable detection granularity inconsistencies caused by generating frame-by-frame proposal masks in SAM everything mode.

\subsection{Robust Object Association}

\noindent\textbf{Agglomerated Encoder.}
We replace the SAM encoder with an agglomerated encoder, RADIO~\cite{ranzinger2023radio}, distilled from multiple vision foundation models including SAM, DINOv2, and CLIP, to provide strong features that have both powerful open-world segmentation capability (as in SAM), and rich visual information for feature association (as in DINOv2).
This design allows VideoSAM to fully exploit and the zero-shot segmentation and fine-grained matching ability of visual foundation model efficiently.

\noindent\textbf{Cycle-ack Pairs Propagation.}
VideoSAM associates objects across frames by tracking position prompts for the mask decoder. Specifically, given the image patch embeddings \( P \in \mathbb{R}^{H_r \times W_r \times d} \) and object masks \( M \in \mathbb{O}^{N \times H_r \times W_r} \) from the reference frame, along with the image patch embeddings from the current frame \( Q \in \mathbb{R}^{H_t \times W_t \times d} \), we aim to find the corresponding position for each object in the current frame to serve as the position prompt.
This process ensures that the segmentation masks are predicted in the order corresponding to their associated position prompts. Unlike the tracking-by-detection paradigm, this method eliminates inefficient frame-by-frame proposals and the need for complex post-processing.
We define a pair of patches from the reference and current frames, \( \mathrm{CP}\{p_i, q_j\} \), as the Cycle-ack Pair if, for a patch \( p_i \in \mathbb{R}^{d} \) in the reference frame, its most similar patch in the current frame is \( q_j \in \mathbb{R}^{d} \), and vice versa, the most relevant patch for \( q_j \) in the reference frame is \( p_i \). Formally, the condition for a Cycle-ack Pair is:
\[ \mathrm{CP}\{p_i, q_j\} = \mathrm{True} ~\text{if}~ q_j = q_j^* ~\text{and}~ p_i = p_i^*, \]
\[ q_j^* = \arg \max_{q_j \in Q} \mathrm{cos}(p_i, q_j), ~ ~ 
p_i^* = \arg \max_{p_i \in P} \mathrm{cos}(q_j, p_i). \]
\( \mathrm{cos}(\cdot, \cdot) \) denotes the cosine similarity, which ranges from \([-1, 1]\). Cycle-ack pairs establish a highly reliable matching relationship, which can then be used to generate accurate position prompts for object masks in the current frame. For each object, as long as we identify the Cycle-ack Pairs from the patches in its mask, we can track the object position in the current frame and encode it as a point prompt for segmentation.
To efficiently identify cycle-ack pairs across all objects, we first compute a global similarity matrix between the reference and current frame features, \( S = \mathrm{cos}(P, Q) \in \mathbb{R}^{H_r W_r \times H_t W_t} \). We then identify all matching pairs through matrix operations:
\[ \arg \max (S, ~\mathrm{d}=H_r W_r) = \arg \max(S, ~\mathrm{d}=H_t W_t), \]
where \( \arg \max (S, ~\mathrm{d}=H_r W_r) \) refers to the matrix argmax operation along the \( H_r W_r \) dimension, yielding indices \( \in \mathbb{N}^{H_t W_t} \).
Next, we verify whether the position of \( q \) falls within the object mask in the reference frame, thus obtaining a set of cycle-ack points for each object in the current frame. For each point set, we select \( K \) points to encode the position prompt. The selection process involves calculating the centrality measure for each point, which is defined as the mean distance to all other points in the same set. After this, a top-k minimum selection is performed to choose the most central points.

\noindent\textbf{Memory Mechanism.}
To enable long-term object tracking, especially when the object undergoes significant deformation, we extend the reference of cycle-ack pairs propagation by introducing the CPs memory. The memory starts with the patch embeddings and object masks from the initial frame. As the video progresses, after each frame object is tracked and segmented, we add the patch embeddings and corresponding object mask (represented as one-hot vectors) of CPs into the memory, provided the point is indeed within the predicted mask.
If the memory reaches its maximum capacity of  $M$  patches, some patches need to be discarded. To manage this, we track the utilization frequency of each CP, which reflects how often each patch has been used as a reference in forming CPs. Patches from the initial frame are preserved since they are the most reliable, and patches from the most recent frame are retained because they likely contain clues that are highly relevant to the current frame. For the remaining patches, we discard those with the lowest CPs utilization rate to make space for new patches, ensuring the memory does not exceed its maximum capacity.

\subsection{Granularity Consistent Segmentation}
\noindent\textbf{AR-SAM Decoder with Object Prompt.}
To address the issue of granularity inconsistency difficult tracking, we introduce AR-SAM Decoder with a temporal autoregressive object prompt. This object prompt is designed to carry forward object-level information from one frame to the next, ensuring that objects are segmented with consistent granularity across the entire video. 
Specifically, in mask decoder of VideoSAM, in addition to the position prompt, used to indicate object positions, and the learnable embeddings of mask tokens for object segmentation, we introduce an object prompt. 
In addition, we insert an extra learned mask token into the set of prompt embeddings. This token is used at the decoder’s output to decode the mask with the specified granularity. A new self-attention layer is also added between all these prompt tokens, positioned between the original two-way attention block in the mask decoder.
The autoregressive object prompt mechanism effectively acts as a video-level granularity-consistent prompt, informing the mask decoder about the expected granularity of objects as they appear in successive frames.

\noindent\textbf{Cycle-sequential Training.}
To training the AR-SAM decoder and further encourage consistent granularity segmentation across frames, we introduce a cycle-sequential training strategy. 
In cycle-sequential training, the network is trained on sequences of consecutive frames during each single iteration.
When segmenting the first frame, the model uses annotated or sampled points and ground-truth masks for training. For subsequent frames, it relies on the predictions from previous frames, utilizing both the maintaining memory and the AR-token to tracking object and maintain consistency.
Moreover, at the end of each sequence, the last frame is cycled back to the first frame. This requires that the AR-token produced by the last frame also generates a mask with the same granularity as the original frame, ensuring consistent segmentation throughout the sequence.
This training regime encourages the model to produce similar segmentations when tracking objects forward and backward through video sequences. 

\section{Experiment}

\begin{table*}[t]
\centering
\caption{Video segmentation performance on UVO~\cite{wang2021unidentified}. Higher is better. Best performing method is bolded.}
\vspace{-1mm}
\label{tab:uvo}
\setlength\tabcolsep{3.6pt}
\begin{tabular}{c|ccccccccc}
\toprule
Method & TAM\cite{yang2023track} & SAM-PT\cite{rajivc2023segment} & SAM-PD\cite{zhou2024sam} & MASA-B\cite{li2024matching} & MASA-H\cite{li2024matching} & DEVA-SAM\cite{cheng2023tracking}  & DEVA-SAM-o\cite{cheng2023tracking}  & OVTrackor~\cite{chu2024zero} & VideoSAM \\
\scriptsize\textcolor{gray}{Source} & \scriptsize\textcolor{gray}{arXiv'2023} & \scriptsize\textcolor{gray}{arXiv'2023} & \scriptsize\textcolor{gray}{arXiv'2024} & \scriptsize\textcolor{gray}{CVPR'2024} & \scriptsize\textcolor{gray}{CVPR'2024} & \scriptsize\textcolor{gray}{ICCV'2023} & \scriptsize\textcolor{gray}{ICCV'2023}  & \scriptsize\textcolor{gray}{ICRA'2024} & \scriptsize\textcolor{gray}{Ours}     \\
\midrule
mAP    & 1.7        & 6.8        & 6.4        & -         & -         & 3.2       & 8.0          & -         & \textbf{9.7}      \\
AR100  & 24.1       & 28.8       & 24.9      & 28.1       & 28.4      & 18.5      & 22.8       & 28.1      & \textbf{29.8}    \\
\bottomrule
\end{tabular}
\vspace{-2mm}
\end{table*}

\begin{figure*}[t]
  \centering
  \includegraphics[width=0.98\linewidth]{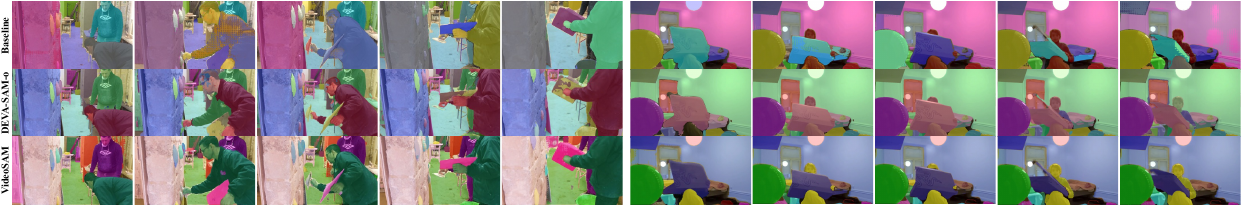}
  \vspace{-1mm}
  \caption{Qualitative results of VideoSAM compared with the SAM baseline and DEVA\cite{cheng2023tracking} on UVO~\cite{wang2021unidentified} and BURST~\cite{athar2023burst} datasets. The baseline prompts SAM with points propagated by feature similarity. In overall, VideoSAM reliably tracks objects and generates object masks with consistent granularity.}
  \label{fig:compare}
    \vspace{-4mm}
\end{figure*}

\subsection{Implementation Details}
\noindent\textbf{Training.} The encoder of VideoSAM is pretrained from RADIO, and frozen during training.  The mask decoder is pretrained from SAM and fine-tuned via LoRA~\cite{hu2021lora}. The parameters of the newly introduced learnable embeddings and layers for object prompts and the autoregressive mechanism are trained from scratch. 
For training, we collected over 5,000 videos with segmentation annotations from video instance segmentation datasets, including YouTubeVIS~\cite{yang2019video}, UVO~\cite{wang2021unidentified}, and BURST~\cite{athar2023burst}. We trained the model for 120k iterations using 8 A10 GPUs, starting with an initial learning rate of 0.001, which was halved at [15k, 30k, 50k, 70k, 90k] iterations. Following SAM, we utilize focal loss and dice loss for mask segmentation, and mean squared error loss for IoU prediction in the mask decoder. 

\noindent\textbf{Inference.} 
VideoSAM uses grid point traversal to detect objects in the first frame, similar to the “everything/automatic” mode in SAM. For the first frame, the object prompt is initially a learnable embedding, and the point prompts come from  a 32x32 grid. We generate N confident object masks after filtering by the predicted IoU, mask stability scores, and NMS. 
For subsequent frames, the position prompts are generated by Cycle-ack Pairs Propagation, and the object prompt is transformed from the previous frame autoregressive token.
For the object-in logic, we perform re-detection every few frames, as the same in DEVA~\cite{cheng2023tracking}.
New objects are identified by matching their masks with the propagated object masks via IoU, and later added to the tracklets. For the object-out logic, we firstly pool N object tokens from the predicted object masks and N tracklet tokens from the historical masks. If the most similar counterpart of an object is not the matched by its own tracklet, we consider that object as “out” for that frame. If an object is out for 10 consecutive frames, it is considered to be permanently out of the video.

\subsection{Benchmarks}
UVO~\cite{wang2021unidentified} is a comprehensive benchmark for open-world, class-agnostic object segmentation in videos, featuring exhaustive annotations. On average, the UVO dataset provides 13.52 annotated instances per video. In contrast, the close-set video instance segmentation benchmark, YouTubeVIS, contains only 1.68 objects per video. This makes UVO a more suitable dataset for evaluating open-world capabilities.

BURST~\cite{athar2023burst} is an open-vocabulary video segmentation benchmark whose sequences are longer (even thousands of frames) compared to UVO. 
In this paper, we evaluate methods on BURST videos in a low-frame-rate scenario (1 FPS), where the tracking becomes more difficult due to the dramatic appearance change. T
Since BURST does not provide annations for all the objects shown in the video, we only evaluate the annotated objects that persist throughout the whole video.


RoboTAP~\cite{vecerik2024robotap} is a point-tracking benchmark for robotic scenarios, providing point annotations but no segmentation mask annotations. We conducted qualitative experiments using robotic operation videos from RoboTAP to preliminarily validate the applicability of the proposed model in real-world robotic vision systems, such as in robot manipulation.

Following \cite{yang2019video, wang2021unidentified}, we adopt  two evaluation metrics: video-level Average Precision (AP) and Average Recall (AR) in the class-agnostic manner, which are commonly used across various datasets and evaluation settings.

\subsection{Evaluation}
\label{sec:eval}
Tab.~\ref{tab:uvo} and Tab.~\ref{tab:burst} quantitatively compare the open-world video segmentation performance of different methods on the UVO and BUSRT benchmarks, respectively. As we can observe, VideoSAM achieves the state-of-the-art performance across almost all metrics, showing its superiority on consistent object tracking and segmentation in both high-frame-rate and low-frame-rate scenarios.
DEVA~\cite{cheng2023tracking}, the previous method with the high AP, 
integrates SAM with a mask propagation module, XMem~\cite{cheng2022xmem}. However, it greatly suffers from the granularity issue from SAM, even using the object-level granularity for inference. DEVA-SAM uses SAM to automatically detect objects at all three levels of granularity, following the default settings of the original SAM. In contrast, DEVA-SAM-o exclusively uses object-level granularity for detection.

\begin{figure}[]
  \centering
  \includegraphics[width=1.0\linewidth]{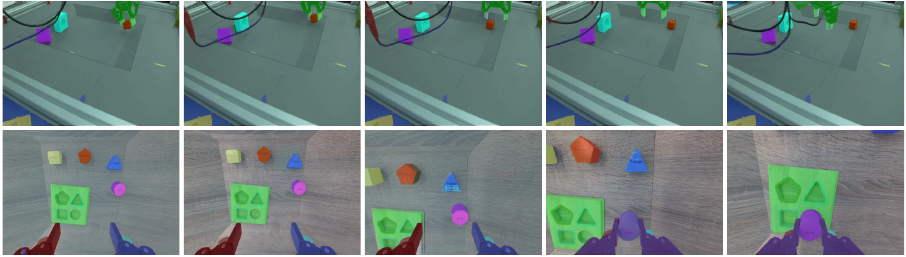}
  \caption{Qualitative performance on the RoboTAP~\cite{vecerik2024robotap} dataset.}
  \label{fig:robo}
  \vspace{-4mm}
\end{figure}

Fig.~\ref{fig:compare} provides a qualitative comparison between VideoSAM and two other methods on UVO and BURST. The baseline uses SAM (with object-level granularity) and follows the position prompt propagation paradigm (by taking the argmax of feature similarity, rather than our cycle-ack pairs propagation) to track objects. 
As shown, VideoSAM consistently tracks and segments objects more stably, while DEVA’s pixel-level mask propagation model occasionally makes localized errors. Compared to the SAM baseline, VideoSAM maintains consistent granularity across frames as expected. Fig.~\ref{fig:robo}, further shows results of VideoSAM on the robotic grasping videos at 3FPS from RoboTAP. Without any finetuning, VideoSAM still produce reliable panoptic segmentation in real-world robotic vision systems.

\begin{table}[]
\centering
\caption{Video segmentation performance on BURST~\cite{athar2023burst}. Higher is better. Best performing method is bolded.}
\label{tab:burst}
\setlength\tabcolsep{2.8pt}
\begin{tabular}{c|c|cccccc}
\toprule
Method             & Source  & mAP           & AP50          & AP75          & AR100         & AR10          & AR1           \\
\midrule
M2F+STCN~\cite{athar2023burst}   & \scriptsize\textcolor{gray}{WACV'23}  & 4.3             & 8.0             & 3.9             & 47.8             & 41.9             &  14.5             \\
DEVA-M2F~\cite{cheng2023tracking}   & \scriptsize\textcolor{gray}{ICCV'23}  & 4.1           & 6.2           & 4.3           & 53.7          & 40.9          & 12.2          \\
DEVA-Entity~\cite{cheng2023tracking} & \scriptsize\textcolor{gray}{ICCV'23} & 6.0             & 8.6           & 6.4           & 52.2          & 41.9          & 14.8          \\
DEVA-SAM~\cite{cheng2023tracking}           & \scriptsize\textcolor{gray}{ICCV'23}  & 4.1           & 8.6           & 3.4           & 37.4          & 27.8          & 6.8           \\
DEVA-SAM-o~\cite{cheng2023tracking}         & \scriptsize\textcolor{gray}{ICCV'23}  & 7.6           & 11.3          & 8.0           & \textbf{54.5} & 43.3          & 8.9           \\
VideoSAM           & \scriptsize\textcolor{gray}{Ours}     & \textbf{10.7} & \textbf{14.9} & \textbf{11.2} & 53.5          & \textbf{45.8} & \textbf{18.1} \\
\bottomrule
\end{tabular}
\vspace{-2mm}
\end{table}

\begin{table}[]
\vspace{-2mm}
\centering
\caption{Ablation study on tracking related components.}
\label{tab:ablt_track}
\begin{tabular}{ccc|cc|cc}
\toprule
\multirow{2}{*}{RADIO} & \multirow{2}{*}{CPs} & \multirow{2}{*}{Mem} & \multicolumn{2}{c}{UVO} & \multicolumn{2}{c}{BURST} \\
                       &                      &                      & mAP        & AR100      & mAP         & AR100       \\
\midrule
\checkmark             & \checkmark           & \checkmark           & 12.1       & 26.6       & 10.7        & 53.5        \\
\texttimes            & \checkmark           & \checkmark           & 9.7        & 28.4       & 10.8        & 47.5        \\
\checkmark             & \texttimes           & \checkmark           & 9.1        & 26.6       & 8.5         & 41.7        \\
\checkmark             & \checkmark           & \texttimes           & 9.2        & 17.1       & 11.0          & 38.8        \\
\texttimes              & \texttimes           & \texttimes           & 7.6        & 15.4       & 7.5         & 29.3       \\
\bottomrule
\end{tabular}
\end{table}

\begin{table}[]
\vspace{-2mm}
\centering
\caption{Ablation study on the points prompt number.}
\label{tab:ablt_pt}
\begin{tabular}{c|ccccc}
\toprule
K & 1    & 2    & 3    & 4   & 5   \\
\midrule
mAP         & 12.1 & 10.4 & 10.1 & 9.9 & 9.9  \\
\bottomrule
\end{tabular}
\end{table}

\begin{figure}[]
  \includegraphics[width=0.96\linewidth]{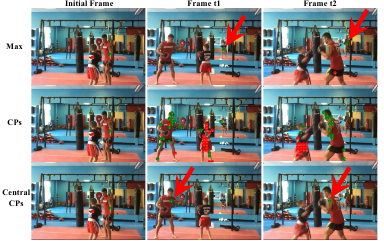}
  \caption{Effect of Cycle-pairs Propagation. Red and green points represent propagated CPs, with red arrows highlighting key results. Best viewed when zoomed in.}
  \label{fig:cps}
  \vspace{-1mm}
\end{figure}

\begin{table}[]
\centering
\caption{Ablation study of granularity related components.}
\label{tab:ablt_seg}
\begin{tabular}{cc|cc|cc}
\toprule
           &            & \multicolumn{2}{c}{UVO}           & \multicolumn{2}{c}{BURST} \\
Obj-Prompt         & Cycle Train   & mAP  & AR100 & mAP         & AR100      \\
\midrule
\checkmark & \checkmark & 12.1 & 26.6  & 10.7        & 53.5       \\
\checkmark & \texttimes & 11.9 & 26.5  & 11.0          & 49.4       \\
\texttimes & \texttimes & 9.2  & 17.1  & 10.9        & 38.8       \\
\bottomrule
\end{tabular}
\vspace{-3mm}
\end{table}

\subsection{Ablation Study}

\noindent\textbf{Object Association.} 
We extensively evaluate key designs of the position-prompt for robust object association and demonstrate their effectiveness. As shown in Tab.~\ref{tab:ablt_track}, the first row presents the default design of the proposed VideoSAM. The row-2 shows that RADIO features improve propagation compared to SAM features. The row-3 demonstrates the significant improvement of Cycle-Pairs over simply using the point with the highest similarity from the correlation map. Fig.~\ref{fig:cps} illustrates the comparison between max correlation and CPs, where CPs offer more reliable propagation due to their strict bidirectional confirmation mechanism. Central CPs further refine this by identifying the most reliable central point. Since our mask decoder with object prompts effectively maintains object granularity, a single point is sufficient as the position prompt to accurately locate the object, as shown in Tab.~\ref{tab:ablt_pt}. The row-4 highlights how continuously updating memory through CPs significantly enhances tracking performance. The final row presents the baseline described in Sec.~\ref{sec:eval}.

\noindent\textbf{Granularity Consistency.}
As shown in Tab.~\ref{tab:ablt_seg}, the object prompt effectively addresses SAM’s granularity inconsistency issue, while cycle-sequential training further enhances the mask decoder’s ability to maintain granularity with the object prompt. In Fig.~\ref{fig:compare}, the second frame of the video on the left shows that VideoSAM preserves the complete granularity of the person segmentation, whereas both the SAM baseline and DEVA exhibit part-errors or over-segmentation. In the video on the right, both SAM and DEVA encounter over-segmentation or under-segmentation issues for the person and the large box, while VideoSAM resolves these problems through the use of object prompts.

\section{Conclusion}
In this paper, we addressed two key challenges in extending SAM to the open-world video segmentation task by proposing VideoSAM, an end-to-end framework. The introduction of Cycle-ack Pairs Propagation with a memory mechanism enables efficient, robust, and stable object tracking across frames. Additionally, the AR-SAM decoder, which incorporates an autoregressive object prompt within the mask decoder, combined with a cycle-sequential training strategy, effectively resolves the issue of inconsistent segmentation granularity through differentiable optimization across frames. Extensive experiments on open-world benchmarks demonstrate the general applicability of VideoSAM, highlighting its potential for real-world robotic applications. 
All code will be available to support the research community. In the future, we plan to explore more complex scenarios involving more frequent object entry and exit as well as camera transitions. Additionally, we will investigate how VideoSAM can be applied to robotic vision systems to improve navigation or manipulation performance.

\clearpage

\bibliographystyle{IEEEtran}
\bibliography{root}

\end{document}